\pdfoutput=1
\documentclass[lettersize,twoside,journal]{IEEEtran}
\usepackage{amsmath,amsfonts}
\usepackage{algorithmic}
\usepackage{algorithm}
\usepackage{array}
\usepackage[caption=false,font=normalsize,labelfont=sf,textfont=sf]{subfig}
\usepackage{textcomp}
\usepackage{stfloats}
\usepackage{url}
\usepackage{verbatim}
\usepackage{graphicx}
\usepackage{cite}
\usepackage{mathrsfs}
\usepackage{color}
\usepackage{colortbl}
\definecolor{Ocean}{RGB}{129,154,254}
\definecolor{orange}{RGB}{254,144,95}
\usepackage{bbding}
\usepackage{pifont}
\usepackage{tabularx}
\usepackage{multirow}
\usepackage{diagbox}
\usepackage{xcolor}
\usepackage[normalem]{ulem}
\usepackage{fancyhdr} 
\fancypagestyle{firstpagefooter}{
    \fancyhf{} 
}

\begin{document}

\title{SGFormer: Spherical Geometry Transformer\\ for 360 Depth Estimation}

\author{ Junsong~Zhang, Zisong~Chen, Chunyu~Lin,
Zhijie~Shen, Lang Nie,
Kang~Liao, Junda~Huang,  Yao~Zhao,~\IEEEmembership{Fellow,~IEEE}
\thanks{}
\thanks{}}

\markboth
{ZHANG \MakeLowercase{\textit{et al.}}: SGFormer: Spherical Geometry Transformer for 360 Depth Estimation}
{}%


\maketitle

\thispagestyle{firstpagefooter} 

\begin{abstract}
Panoramic distortion poses a significant challenge in 360 depth estimation, particularly pronounced at the north and south poles. Existing methods either adopt a bi-projection fusion strategy to remove distortions or model long-range dependencies to capture global structures, resulting in either unclear structure or insufficient local perception. 
In this paper, we propose a spherical geometry transformer, named \textbf{SGFormer}, to address the above issues, with an innovative step to integrate spherical geometric priors into vision transformers.
To this end, we retarget the transformer decoder to a spherical prior decoder (termed \textbf{SPDecoder}), which endeavors to uphold the integrity of spherical structures during decoding. Concretely, we leverage bipolar reprojection, circular rotation, and curve local embedding to preserve the spherical characteristics of equidistortion, continuity, and surface distance, respectively.
Furthermore, we present a query-based global conditional position embedding to compensate for spatial structure at varying resolutions. It not only boosts the global perception of spatial position but also sharpens the depth structure across different patches.
Finally, we conduct extensive experiments on popular benchmarks, demonstrating our superiority over state-of-the-art solutions. Our code will be made publicly at \textcolor{magenta}{\url{https://github.com/iuiuJaon/SGFormer}}.
\end{abstract}

\begin{IEEEkeywords}
Depth Estimation, Panoramic Distortion, Spherical Priors, Position Embedding
\end{IEEEkeywords}

\section{Introduction}
\begin{figure}[ht]
\centering
\includegraphics[scale=0.19]{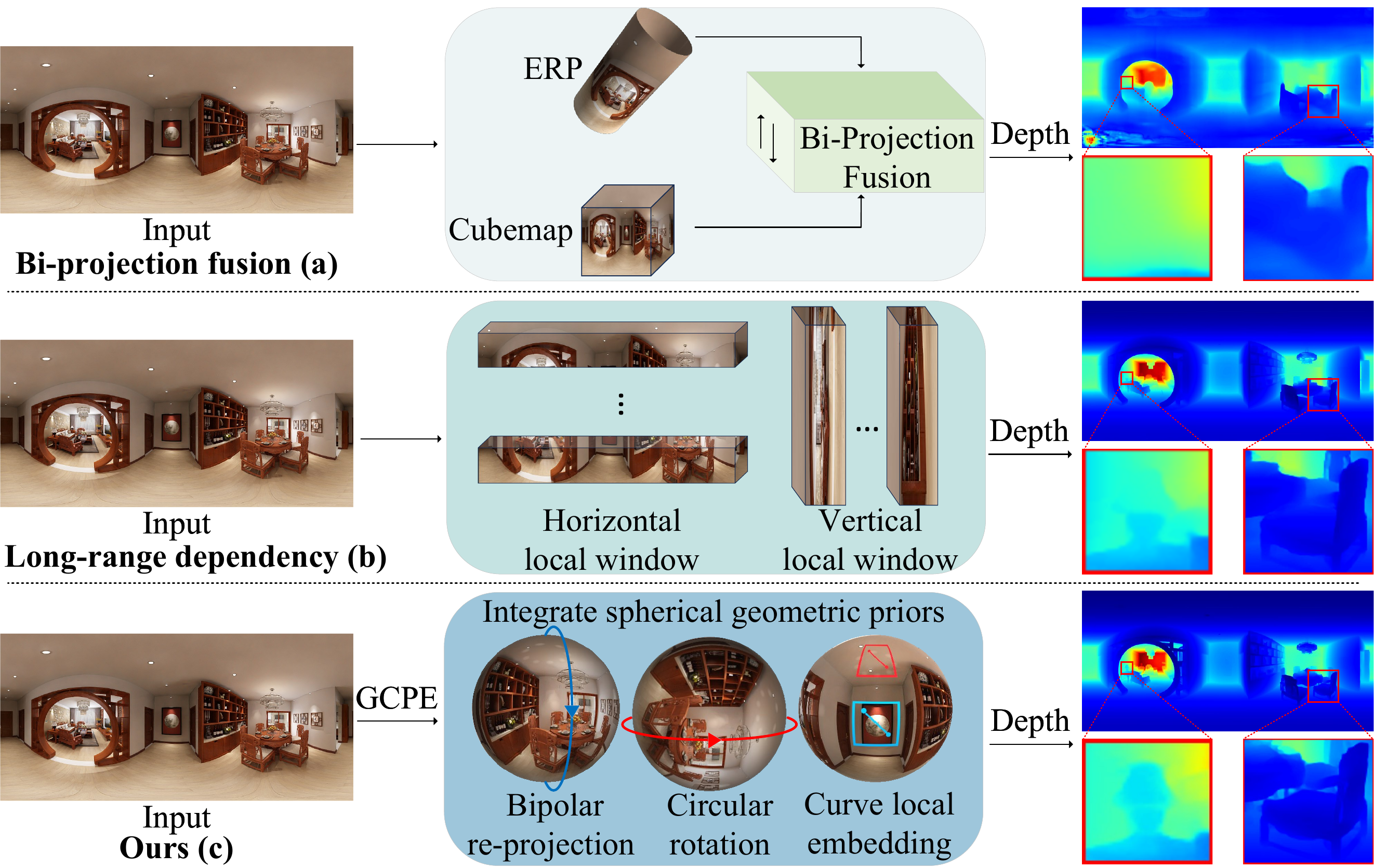}
\caption{\textbf{Brief comparisons between previous typical methods and the proposed method: (a) Bi-projection fusion strategy. Bifuse \cite{r1} combines feature maps from equirectangular and cubemap projections. (b) Long-range dependency strategy. EGFormer \cite{r7} adapts vertical and horizontal shaped local windows. (c) Our SGFormer presents a query-based global conditional position embedding (GCPE) and integrates spherical geometric priors, significantly improving the performance of panoramic depth estimation.}}
\label{fig1}
\end{figure}

The panoramas represent a whole scene as an equirectangular map, delivering a 360 Field of View (FoV) of the surroundings. Hence, estimating depth from a single panoramic image can offer complete depth of the scene, with a holistic understanding of the surroundings. However, the equidistant unfolding characteristic (also known as panoramic distortion) of panoramas poses a new challenge for 360 depth estimation. Therefore, directly applying the perspective-based approaches \cite{r46,r47,r48,r49,r50,r51,r53} to estimate depth from a single panorama is challenging. To address this issue, researchers \cite{r1,r4} have recently preferred to estimate accurate panoramic depth with the assistance of other projection formats (\textit{e.g.}, Cubemap projection (CP), and Tangent projection (TP)). 
These projections significantly reduce distortions by mapping a panorama onto several tangent faces but damage the inherent continuous representation of panoramic images.
Many approaches \cite{r1,r2,r3,r4} typically leverage the complementarity of these projection representations (shown in Fig. \ref{fig1}a) to estimate more accurate panoramic depth. Nevertheless, the trade-off between the FoV continuity and distortion elimination poses another challenge for the model design. Besides, these bi-projection fusion strategies require additional branches, increasing the computational burden. In view of this, we only take the equirectangular projection as the input and delve into the properties of the spherical space to deal with the distortion.

On the other hand, some approaches \cite{r5,r6,r7} deal with the panoramic distortions via modeling long-range dependencies (as illustrated in Fig. \ref{fig1}b). In previous work \cite{r5}, long short-term memory \cite{r36} (LSTM) is introduced to disseminate global information in vertically compressed sequences. However, the geometric structure, which is an essential cue to estimate depth, is neglected, thus delivering blurred depth. Meanwhile, those methods based on modeling long-range dependencies are either too global to perceive the local details \cite{r7} or too local to deal with the distortion efficiently \cite{r6}.

In this paper, we incorporate spherical geometry into our model (\textit{i.e.}, SGFormer, as shown in Fig. \ref{fig1}c) to capture local details while promoting global interaction. 
We leverage three geometric priors of the sphere to design a customized panoramic decoder (SPDecoder). 
First, we use the spherical equidistortion to design a bipolar re-projection paradigm to deal with the distortions. We re-project these areas with severe distortions to the equator region to eliminate the negative effect of distortions. Specifically, our method is parameter-free and transformation-efficient without any other branches.
Subsequently, we perform circular rotation through the continuity of the sphere.
It helps to produce seamless panoramic representations, which allows two separate points, originally located on the left and right boundaries, to be close to each other and have similar semantics.
Furthermore, with the spherical surface distance, curve local embedding is presented to generate curved relative position embedding for each point in the spherical projection space, which provides additional geometric distance and enhances the understanding of spatial structure.

To enhance performance comprehensively, we further design a query-based global conditional position embedding (GCPE) scheme. By leveraging features of different resolutions as queries to query the keys that match the resolution, we can adaptively provide different geometric details and spatial structures at different resolutions.
This additional structure, in turn, explicitly guides our SPDecoder with global spatial positioning, thereby improving the accuracy of estimated depth.

We evaluate the proposed SGFormer on two benchmarks, Structured3D \cite{r9} and Pano3D \cite{r10}. Experimental results demonstrate that our method outperforms existing methods by a large margin on quantitative and qualitative results. Our main contributions can be summarized as follows:

\begin{itemize}
\item We propose a spherical geometry transformer, named SGFormer, to integrate spherical structures and global 
 positional perception simultaneously, contributing to superior performance over state-of-the-art (SoTA) solutions. 
\item We design the SPDecoder by incorporating three spherical geometric priors: equidistortion, continuity, and surface distance, preserving the structural integrity of the sphere and enhancing the perception of spherical geometry and local details.
\item We introduce a query-based global conditional position embedding, which provides an explicit geometric cue, thereby sharpening the depth structure across various patches. 
\end{itemize}

\section{Related Work}
\subsection{Monocular 360 Depth Estimation}
\textbf{Bi-projection fusion strategy.}
BiFuse \cite{r1} introduces a dual-branch fusion method that combines CP and ERP features to alleviate distortion, while UniFuse \cite{r2} proposes a new framework that unidirectionally fuses CP features into ERP features during the decoding stage, enhancing both efficiency and performance.
Glpanodepth \cite{r11} proposes a fusion module that combines ERP and CP features for panoramic depth estimation.
360MonoDepth \cite{r12} and \cite{r67} introduce the framework that uses tangent images for depth estimation from high-resolution 360 images.
OmniFusion \cite{r3} directly converts 360 images into less-distorted perspective patches and uses geometric priors to merge the predictions of these patches into the ERP space to produce the final depth output in ERP format. 
HRDFuse \cite{r4} employs a collaborative mechanism that explores both the overall contextual information of the ERP and the local structure of the TP, estimating panoramic depth by learning the similarity of features between the ERP and TP.

\textbf{Modeling long-range dependencies.}
SliceNet \cite{r5} divides the ERP image into vertical slices. However, it is hard to recover the details near the poles because it ignores the latitudinal distortion property. 
PanoFormer \cite{r6} attempts to estimate panoramic depth by leveraging a transformer-based architecture to process TP patches. By sampling the eight most relevant tokens for each central token on the ERP domain and adaptively adjusting local window sizes with a learnable offset, it enhances the precision in capturing the 2D structure and spatial details of each patch. 
EGFormer \cite{r7} leverages equirectangular geometry as the bias for local windows, thus estimating panoramic depth through an equirectangular geometry-aware local attention with a large receptive field. However, the vertical and horizontal shaped local windows also weaken the perception of the structure of each local area.

\textbf{Other perspectives.}
Some researchers  \cite{r13,r14,r17,r18,r20,r21,r22} employ various distortion-aware convolution filters to mitigate distortion, but due to the limited receptive fields, the final effect is still not entirely satisfactory.
In deeper layers, several spherical CNNs \cite{r16,r23} have employed convolution within the spectral domain, which could lead to a heavier computational burden.
 \cite{r25} considers the significance of the geometric structure in depth estimation and proposes a neural contourlet network for panoramic monocular depth estimation.
Different from supervised approaches,  \cite{r72} proposes an unsupervised domain adaptation strategy that leverages unlabeled 360 data to fine-tune existing indoor panoramic depth estimation models (e.g., BiFuse++ or UniFuse). Specifically, it employs Depth Anything \cite{r75} to generate pseudo-labels for supervision. Hence, its performance is constrained by the base models and the quality of the pseudo-labels. In contrast, we integrate spherical geometric priors into the SPDecoder to design the base model.
\cite{r74} introduces SegFuse, an end-to-end two-branch multi-task learning network designed for 360-depth estimation, which focuses more on outdoor than indoor scenarios.
In driving scenarios, \cite{r70} proposes an innovative framework to address the challenges of imperfect observation and projection fusion in panoramic depth estimation.  \cite{r73} presents the ADUULM-360 dataset, a multimodal dataset for depth estimation in adverse weather, which focuses more on outdoor driving scenarios. In addition, its dataset and code have not been released yet.

\subsection{Transformer for Vision Tasks}
\begin{figure*}[ht]
\centering
\includegraphics[scale=0.4]{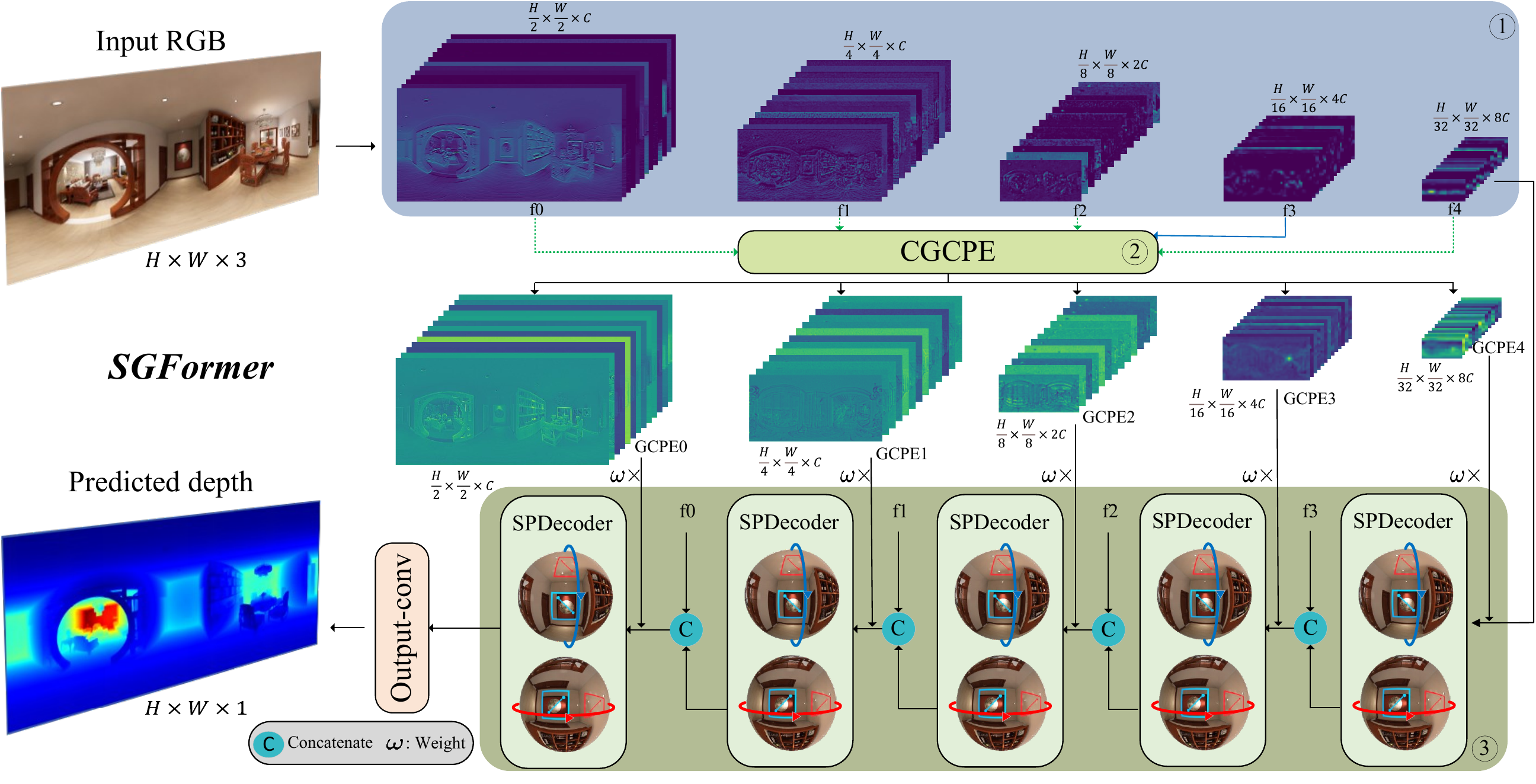}
\caption{\textbf {Overview of our proposed SGFormer. The SGFormer comprises three major parts: feature extractors for ERP images at varying resolutions, calculation of global conditional position embedding (CGCPE) module, and the SPDecoder that integrates three spherical geometric priors. }}
\label{fig2}
\end{figure*}

Inspired by the transformer in the NLP field, the Vision Transformer \cite{r26} (ViT) uses a global self-attention mechanism for visual tasks. 
However, the global information interaction often results in higher computational costs. Swin Transformer \cite{r27} (SwinT) adopts a window-based self-attention, incorporating a hierarchical architecture and shifting window mechanism for image processing. 
\cite{r71} employs a convolution-free swin transformer as an image feature extractor. While SwinT is primarily used as a backbone in the architecture, our contribution is mainly focused on the decoder.
CSwin Transformer \cite{r8} (CSwinT) introduces a cross-shaped window attention mechanism, conducting parallel self-attention computations on horizontal and vertical stripes. 
 \cite{r28,r29} propose distortion relief methods specifically designed for panoramic image vision tasks.
RCDFormer \cite{r68} integrates heterogeneous data from radar and cameras, leveraging a transformer architecture for dense depth estimation.
Despite these advancements, the distortion in panoramic images poses some challenges for transformers in panoramic image processing.
For example, the SwinT does not consider the distortion, equidistortion, and continuity of panoramic images. The SPDecoder is specially designed for panoramic images and can effectively cope with and utilize these unique properties.

PanoFormer \cite{r6} redesigns the traditional transformer with tangent patches on the spherical domain and panorama-specific metrics, using learnable token flows to perceive geometric structures adaptively. 
HRDFuse \cite{r4} processes panoramic images into TP patches and aggregates global information in a ViT manner. 
EGFormer \cite{r7} combines with CSwinT \cite{r8} to propose an equirectangular geometry-biased transformer. However, this horizontal and vertical self-attention weakens the perception of local structures. In this work, we use the spherical geometric characteristics to further explore the suitable transformer structure for panoramic views.

\subsection{360 distortion processing and model lightweighting}
To mitigate the adverse effects of distortion, \cite{r54} proposes a multi-projection fusion and refinement network. Recently, a simple and effective Cube Padding (CP) technique \cite{r57} has been proposed, demonstrating that the CP projection format can benefit panoramic tasks. 
For 360 video saliency detection, \cite{r58} introduces an innovative scheme based on the spherical convolutional neural network.  
DATFormer \cite{r69} addresses distortions with two modules: one for global distortion adaptation and another for reducing local distortions on multi-scale features.
To handle distortions in spherical images, \cite{r59} presents a distortion-aware convolutional network, while \cite{r60} designs a spherical deep neural network (DNN) specifically to overcome distortions caused by the equirectangular projection. 
DINN \cite{r65} proposes a novel deformable invertible neural network (INN) for latitude-aware 360 image rescaling.
New spherical image representations for 360 images are proposed by \cite{r61} and \cite{r62}. Additionally, \cite{r63} and \cite{r64} focus on developing lightweight models.

\section{SGFormer}
\subsection{Architecture Overview}
\label{sec3_1}
\begin{figure}[t]
\centering
\includegraphics[scale=0.34]{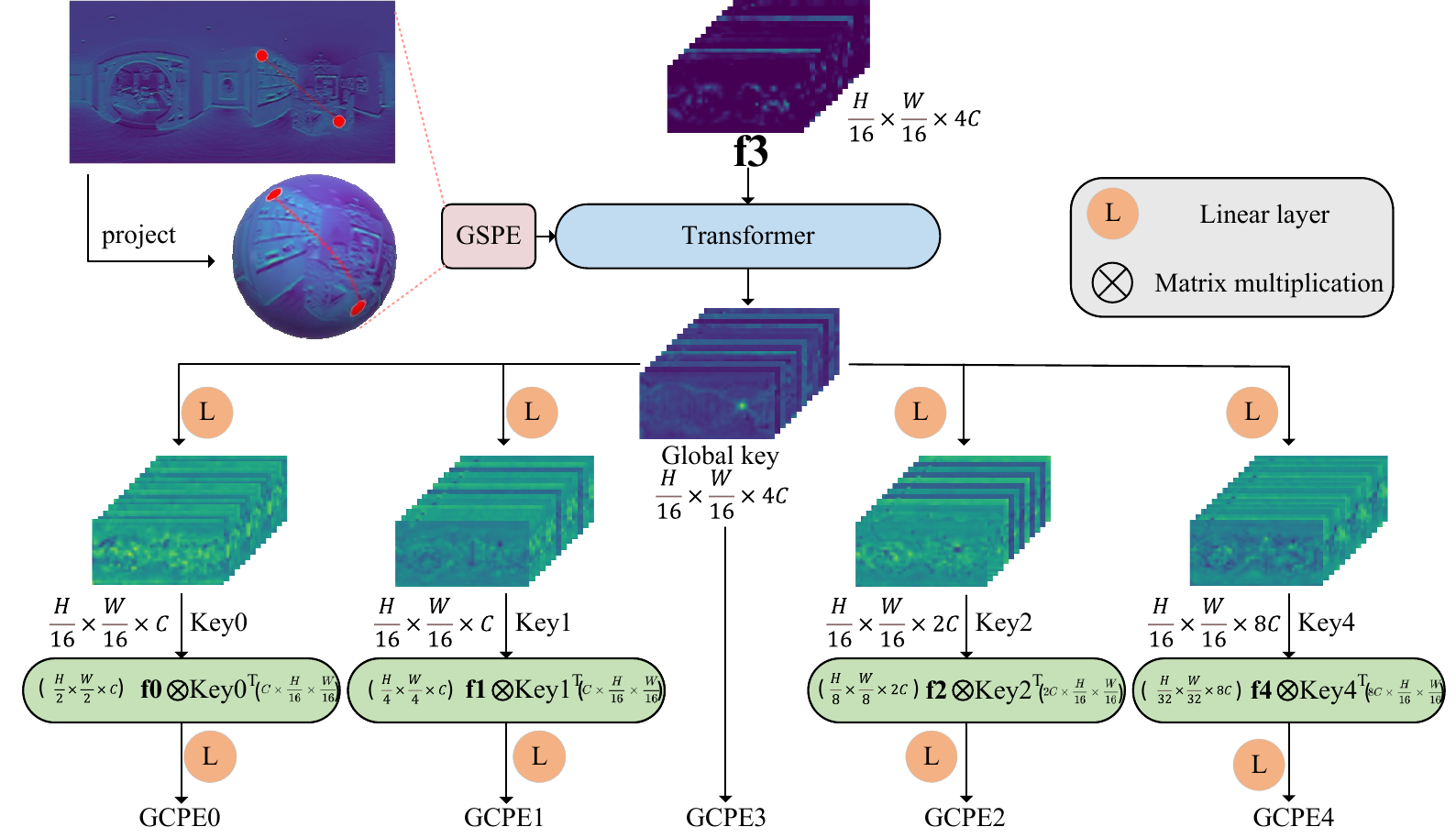}
\caption{\textbf {Overview of the query-based CGCPE module. In particular,  the extracted feature f3 generates a global key (\textit{i.e.} GCPE3) by passing through a transformer module with global spherical position embedding (GSPE). The global key is then processed through linear layers to generate keys corresponding to features at various resolutions. The original features of different resolutions are used as queries, which are subsequently operated by $QK^{T}$. After another linear transformation, the corresponding GCPEs at different resolutions are obtained.}}
\label{fig3}
\end{figure}
In Fig. \ref{fig2}, we show our complete framework motivated by the spherical properties. Overall, the proposed SGFormer takes a panorama as input and outputs a corresponding high-quality depth map. During the encoder stage, we employ Resnet-34 to extract features at five different resolutions (Fig. \ref{fig2} \ding{172}). Then, these features are fed into the calculation of the global conditional position embedding (CGCPE) module (Fig. \ref{fig2} \ding{173} ), which is a query-based module that adaptively produces global conditional position embedding (GCPEs), compensating for additional spatial geometric structure (More details are presented in Section \ref{sec3_2}). The visualization results of GCPEs demonstrate that they provide geometric details, promoting a comprehensive understanding of spatial location. Subsequently, a hierarchical decoder (the SPDecoder, described in Section \ref{sec3_3}) is designed to estimate depth from these extracted features and integrates three spherical geometric priors to maintain the unity of the spherical geometric structure. As benefits by explicit structural cues and spherical priors (\textit{i.e.} equidistortion, continuity, and surface distance), our scheme can effectively relieve distortion, and these structure- and sphere-aware designs can promote the performance of our model. 

\subsection{Calculation of Global Conditional Position Embedding (CGCPE) Module}
\label{sec3_2}
We find that the structure of the scenes is crucial for depth estimation.
Hence, we aim to leverage the panoramic geometric cue to assist in the depth estimation process. To achieve this, we introduce GCPEs, which are learned from the extracted features and a spherical surface distance (\textit{i.e.} Fig. \ref{fig3} global spherical position embedding (GSPE)), to conduct the depth estimation with an additional explicit structure cue. The introduced GCPEs provide scale-adaptive variable details, which can push the network to explore the diverse spatial structures of the same panoramic space but with different resolutions.

Specifically, as shown in Fig. \ref{fig3}, we select the feature with a $\frac{H}{16} \times  \frac{W}{16}$ resolution (denoted as $f_3$) as the reference feature. There is a trade-off between computational efficiency and clear details. To balance them, we choose the middle scale. The selected features are fed into a transformer module. Then we achieve position embedding via a GSPE approach (more details are in section \ref{sec3_3}, curve local embedding), thereby promoting the generation of panoramic structure (\textit{i.e.}, global key in Fig. \ref{fig3}). As shown in the top-left corner of Fig. \ref{fig3}, GSPE projects any two pixels in the ERP feature map into the spherical space to calculate the spherical surface distance as a sphere-aware embedding. Subsequently, the aggregated global geometric information is transformed into keys via a linear projection layer. At each scale, we leverage features as queries to query the keys, adaptively generating GCPE at each resolution.

\begin{figure}[ht]
\centering
\includegraphics[scale=0.55]{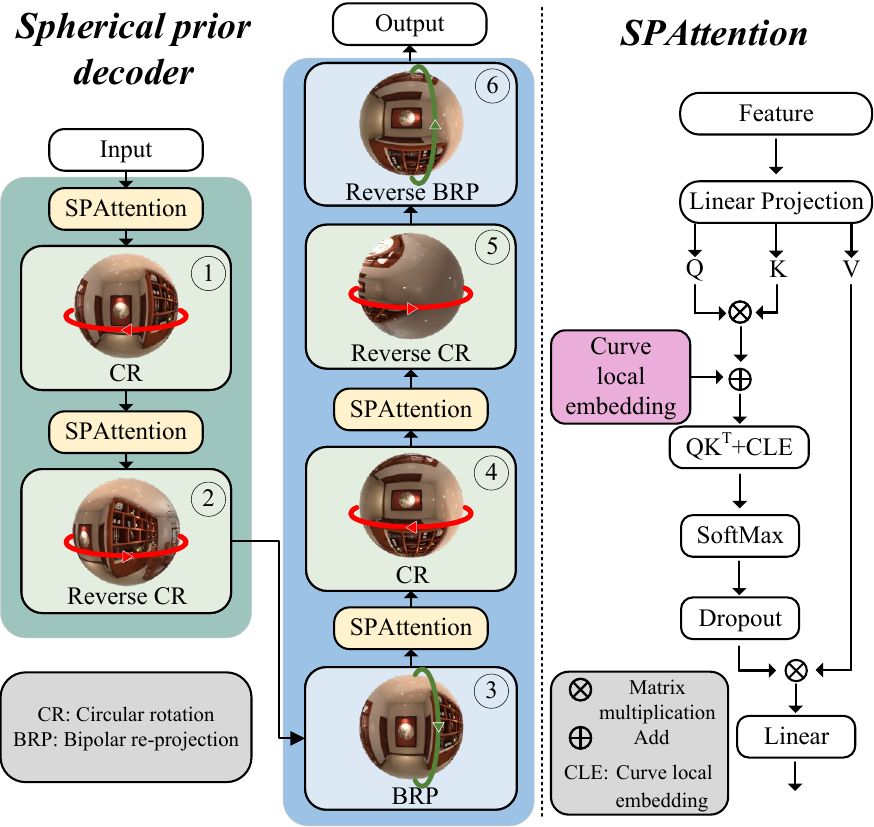}
\caption{\textbf {Overview of SPDecoder (left), which is combined with three spherical geometric priors, and SPAttention (right).}}
\label{fig5}
\end{figure}

\subsection{SPDecoder}
\label{sec3_3}
To balance the global perception and the local details enhancement, we propose the SPDecoder by leveraging spherical priors. As illustrated in Fig. \ref{fig5}, we first send the input features to SPAttention, where we add curve local embedding to provide additional sphere-aware geometric distance. Then, considering the sphere's continuity, we perform circular rotation to enhance the feature interaction among patches (Fig. \ref{fig5} \ding{172}). We restore the original state through reverse circular rotation (Fig. \ref{fig5} \ding{173}). 
Afterward, a bipolar re-projection approach based on the spherical equidistortion property is designed to relieve the distortion (Fig. \ref{fig5} \ding{174}). Moreover, circular rotation is operated again to interact with the original features in the vertical direction (Fig. \ref{fig5} \ding{175}). Ultimately, we revert to the original state through reverse operations (Fig. \ref{fig5} \ding{176}\ding{177}). 
We describe these operations in detail as follows.

\begin{figure}[ht]
\centering
\includegraphics[scale=0.35]{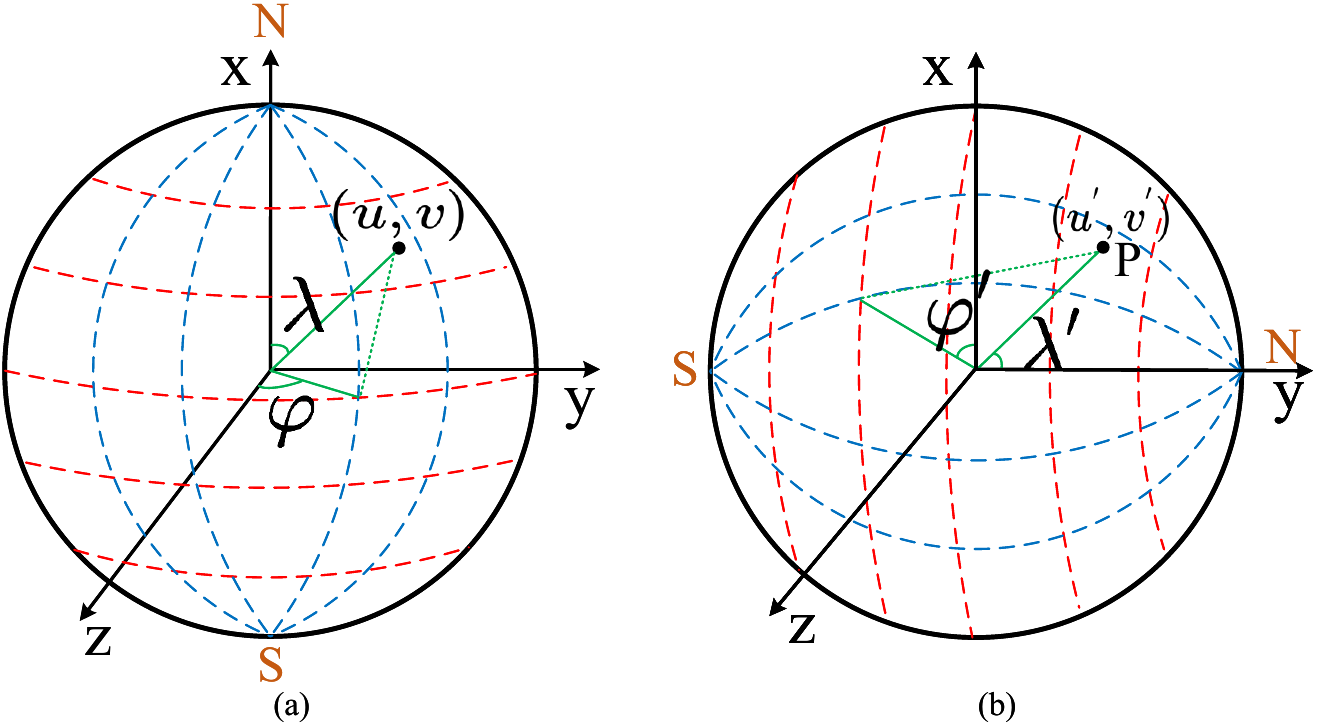}
\caption{\textbf {(a) A projection imaging model with the x-axis (North) as the positive direction. (b) A projection imaging model with the y-axis (North) as the positive direction.}}
\label{fig6}
\end{figure}

\textbf{Bipolar re-projection (BRP):}
Since ERP images exhibit significant distortion in the polar regions and slight distortion near the equator areas, 
BRP can provide a content-consistent but perspective-different panorama via a bipolar transformation operation.
Given a feature map $I \in R^{C\times H\times W}$, we implement BRP through a back projection operation. Specifically, we generate a canvas $I^{\prime } \in R^{C\times H\times W}$ and then transform the canvas from the ERP domain to a unit 3D Cartesian coordinates (Fig. \ref{fig6} b). $\forall P \left ( u^{\prime },v^{\prime } \right ) \in I^{\prime }$, the transformation process can be denoted as follows:
\begin{equation}
\left\{\begin{matrix}\lambda ^{\prime  }=  \frac{v^{\prime}}{H} \times \pi  \\\varphi ^{\prime }=  \frac{u^{\prime}}{W} \times  2\pi \end{matrix}\right.
\label{equ1}
\end{equation}
where $( u^{\prime },v^{\prime })$ indicates the position of the point P in ERP domain; $\lambda ^{\prime } \in \left ( -\frac{\pi }{2} ,\frac{\pi }{2} \right )  $/$\varphi  ^{\prime } \in \left ( -\pi,\pi \right ) $ represents latitude/longitude, respectively. Subsequently, we can calculate the $\left ( x,y,z \right )$  coordinates corresponding to $ \left ( \lambda ^{\prime } ,\varphi ^{\prime } \right )$:
\begin{equation}
\left\{\begin{matrix}x=\sin \left ( \lambda ^{\prime  }  \right )  \\y=\cos \left ( \lambda ^{\prime  }   \right )  \\z=\cos \left ( \frac{\pi }{2}- \lambda ^{\prime  }   \right ) \cdot \sin \left ( \varphi  ^{\prime  }  \right ) \end{matrix}\right.
\label{equ2}
\end{equation}
Similarly, the original feature map points can also be transformed into the same unit 3D Cartesian coordinates. And they have the following mapping relationships:
\begin{equation}
\left\{\begin{matrix}\lambda  =\arccos \left ( \frac{x}{\sqrt{z^{2}+y^{2} } }  \right )\\\varphi =\arcsin \left ( \frac{y}{\sqrt{z^{2}+y^{2} } } \right ) \end{matrix}\right.
\label{equ3}
\end{equation}
where $\lambda \in \left ( -\frac{\pi }{2},\frac{\pi }{2} \right ) $ and $\varphi \in \left ( -\pi,\pi \right ) $ are the spherical coordinates of the original feature map (Fig. \ref{fig6} a). Then we can obtain the position of point $P \left ( u^{\prime },v^{\prime } \right )$ in the original feature map through the following formula:
\begin{equation}
\left ( u ,v\right ) =\left ( \frac{\lambda  }{\pi  } \times H, \frac{\varphi } {2\pi  } \times W\right ) 
\label{equ4}
\end{equation}
In particular, we employ the bilinear interpolation function to fill the missing pixels. After BRP operation, the polar regions are projected to the equator, which allows the model to estimate accurate depths with these distortion-free features. Besides, the proposed BRP approach is parameter-free and efficient in transformation, and no additional branches are required.


\textbf{Circular rotation (CR):}
The traditional transformer \cite{r37} architecture primarily relies on a global self-attention mechanism, which is inefficient when dealing with high-resolution images. To address this, SwinT \cite{r27} performs self-attention among divided window-based patches and employs a shifting window operation to facilitate feature interaction between the adjacent patches. Due to the unique seamless property of panoramic images, we implement self-attention interactions between local windows by rotating half the window size in the horizontal direction. Then, we rotate back to the original state to align with the initial spatial representation. 
Since the north and south poles of panoramic images are not adjacent in space, we cannot directly apply CR in the vertical direction. However, benefiting from the BRP module, we can map the features from the vertical to the horizontal direction. Then, through CR, features that are originally in the vertical direction are interacted. Finally, we obtain a feature representation that combines horizontal and vertical directions.

\begin{figure}[ht]
\centering
\includegraphics[scale=0.7]{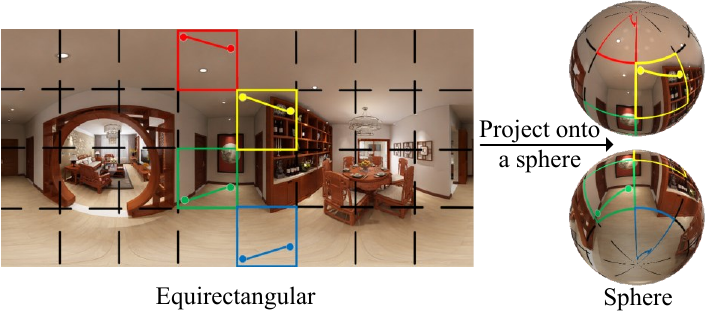}
\caption{\textbf {The left image shows an ERP format panoramic image, with each square indicating a segmented patch. Red and blue squares mark patches from the top and bottom with high distortion, while yellow and green squares indicate middle area patches with less distortion. These patches are projected onto the spherical space on the right.}}
\label{fig8}
\end{figure}
\textbf{Curve local embedding (CLE):}
In the ERP domain, whether in the polar regions with significant distortion or near the equator with less distortion, the distances between two points within a patch are relatively close regardless of the degree of distortion (on the left side of Fig. \ref{fig8}). However, after projecting onto the spherical domain, the distances between these points vary significantly according to the degree of distortion (on the right side of Fig. \ref{fig8}). The network can perceive the panoramic distortion distribution from the mutative distance between the two relevant points.

Hence, we introduce CLE to calculate the distance between two pixels in spherical space. In the SPAttention, we employ CLE as a relative position embedding to model the distortion distribution pattern, thereby delivering an explicit spherical geometry to make the model perceive the distortion easily.

For each patch containing N points, we define $C$ as the CLE within each ${N\times N}$ patch. $\forall P_{1} \left ( \lambda _{1},\varphi  _{1}  \right ) $ and  $P_{2} \left ( \lambda _{2},\varphi  _{2}  \right ) $ (\textit{i.e.}, the position of two points within a patch in spherical coordinates) to calculate the CLE. $c\left ( P_{1}, P_{2}\right ) $ represents the distance between $P_{1}$ and $P_{2}$, where $\lambda \in \left ( - \frac{\pi }{2},\frac{\pi }{2}  \right ) $/$\varphi \in \left ( - \pi,\pi  \right ) $ represents latitude/longitude, respectively. Through the Haversine formula \cite{r39}, we can calculate $c\left ( P_{1}, P_{2} \right ) $ as follows:
\begin{equation}
h=\sin ^{2} \left ( \frac{\lambda _{2}-\lambda _{1} }{2}  \right ) +\cos \left ( \lambda _{1} \right )\cdot \cos \left ( \lambda _{2}  \right )\cdot\sin ^{2} \left ( \frac{\varphi  _{2}-\varphi  _{1} }{2}  \right ) 
\label{equ5}
\end{equation}
\begin{equation}
c\left ( P_{1},P_{2}\right ) =2R\cdot \arctan \left ( \sqrt{\frac{h}{1-h} }  \right ) 
\label{equ6}
\end{equation}
where the radius (R) is set to 1. We employ Eq. \ref{equ5} and  Eq. \ref{equ6} to obtain the CLE within a patch.

\begin{table*}[htbp]\centering
\setlength{\tabcolsep}{9pt}
\caption{ \centering Quantitative comparison on the S3D (Structured3D) and Pano3D datasets with the SoTA methods. {\color{blue}\textbf{Blue}} indicates that our method achieves the best performance.
}
\label{table1}

{
\begin{tabular}{c c c|c c c c|c c c}
\hline
\textbf{Dataset} & \textbf{Method} & \textbf{Pub'Year} &\cellcolor{orange} \textbf{Abs.rel $\downarrow$}  & \cellcolor{orange}\textbf{Sq.rel $\downarrow$} & \cellcolor{orange}\textbf{RMS.lin $\downarrow$}  & \cellcolor{orange}\textbf{RMSlog $\downarrow$} &\cellcolor{Ocean} \textbf{  $\delta ^1$ $\uparrow$} &\cellcolor{Ocean} \textbf{ $\delta^2$ $\uparrow$} & \cellcolor{Ocean}\textbf{  $\delta^3$ $\uparrow$} \\
\hline
            
            & Bifuse  \cite{r1} &CVPR'20    &0.0644 &0.0565 &0.4099 &0.1194 &0.9673 &0.9892 &0.9948 \\
            & SliceNet  \cite{r5} &CVPR'21  & 0.1103 & 0.1273 & 0.6164 & 0.1811 & 0.9012 & 0.9705 & 0.9867\\
            & Yun \textit{et al.}  \cite{r30} &AAAI'22 &0.0505 &0.0499 &0.3475 &0.1150 &0.9700 &0.9896 &0.9947\\
S3D  \cite{r9}  & PanoFormer  \cite{r6} &ECCV'22 &0.0394 &0.0346 &0.2960 &0.1004 &0.9781 &0.9918 &0.9958\\
            & EGFormer  \cite{r7} &ICCV'23   &0.0342 &\textbf{0.0279} &0.2756 &0.0932 &0.9810 &0.9928 &0.9964\\
            & Elite360D  \cite{r45} &CVPR'24 &0.0391 &0.0350 &0.2804 &0.0971 &0.9799 &0.9925 &0.9962\\
            & SGFormer (ours) &-     &{\color{blue}\textbf{0.0303}} &0.0293 &{\color{blue}\textbf{0.2429}}  &{\color{blue}\textbf{0.0863}} &{\color{blue}\textbf{0.9857}} & {\color{blue}\textbf{0.9940}} & {\color{blue}\textbf{0.9969}}\\
\hline  \hline 
            & Bifuse  \cite{r1} &CVPR'20   &0.1704 &0.1528 &0.7272 &0.2466 &0.7680 &0.9251 &0.9731\\
            & SliceNet \cite{r5} &CVPR'21   &0.1254 &0.1035 &0.5761 &0.1898 &0.8575 &0.9640 &0.9867\\
            & Yun \textit{et al.} \cite{r30} &AAAI'22  &0.0907 &0.0658 &0.4701 &0.1502 &0.9131 &0.9792 &0.9924\\
Pano3D  \cite{r10}   & PanoFormer \cite{r6} &ECCV'22  &0.0699 &0.0494 &0.4046 &0.1282 &0.9436 &0.9847 &0.9939\\
            & EGFormer \cite{r7} &ICCV'23 & 0.0660 &0.0428 &0.3874 &0.1194 &0.9503 &0.9877 &0.9952\\
            & Elite360D  \cite{r45} &CVPR'24 &0.0713 &0.0438 &0.3888 &0.1236 &0.9482 &0.9870 &0.9950\\
            & SGFormer (ours) &- & {\color{blue}\textbf{0.0583}} & {\color{blue}\textbf{0.0364}} & {\color{blue}\textbf{0.3537}}  &{\color{blue}\textbf{0.1101}} &{\color{blue}\textbf{0.9613}} & {\color{blue}\textbf{0.9896}} & {\color{blue}\textbf{0.9957}}\\
\hline     \hline     
\end{tabular}
}
\end{table*}

\subsection{Loss Function}
Similar to the existing methods \cite{r7}, the scale-and-shift-invariant loss  \cite{r38} is introduced to eliminate the influence of depth scale and offset differences between different datasets. Specifically, we align the predicted depth $ D_{p}$ and the ground truth depth $D_{g}$ with their scale (s) and shift (t) appropriately through a least-squares criterion:
\begin{equation}
s,t=\arg\min\limits_{s,t}\left ( s\cdot D_{p}+t- D_{g}  \right ) 
\end{equation}
\begin{equation}
\begin{aligned}
D_{p}^{\prime } =s\cdot D_{p}+t    
\end{aligned}
\quad
\begin{aligned}
D_{g}^{\prime }=D_{g}
\end{aligned}
\end{equation}
where $D_{p}^{\prime }$ and $D_{g}^{\prime }$ are the aligned predicted depth and the corresponding ground truth. Subsequently, we calculate the mean squared error loss ($L_{pix}$) as follows:
\begin{equation}
m=\left ( D_{g}^{\prime }   > 0  \right ) 
\end{equation}
\begin{equation}
L_{pix} =\frac{1}{n} \sum_{i=1}^{n} \left [ m\cdot \left | \left ( D_{p}^{\prime } -D_{g}^{\prime }  \right )  \right |  \right ] 
\end{equation}
where the mask (m) is employed to filter out the invalid values in ground truth. Besides, the gradient loss ($L_{grad}$) is employed to sharpen the depth edges, which is calculated as follows:
\begin{equation}
L_{grad} =\sum _{x,y} \left ( \left | G_{x} \left ( D_{p}^{\prime } -D_{g}^{\prime }   \right ) _{x,y}\right  |  +\left | G_{y} \left ( D_{p}^{\prime } -D_{g}^{\prime }  \right ) _{x,y}  \right  |  \right ) 
\end{equation}
where $G_{x}$ and  $G_{y}$ are the horizontal/vertical gradients of a predicted depth map. Ultimately, the total objective function can be written as:
\begin{equation}
L_{total} =L_{pix}+\omega \cdot L_{grad}
\end{equation}
we empirically choose $\omega$ as 0.5.
\vspace{5pt}


\begin{figure*}[t]
\centering
\includegraphics[scale=0.45]{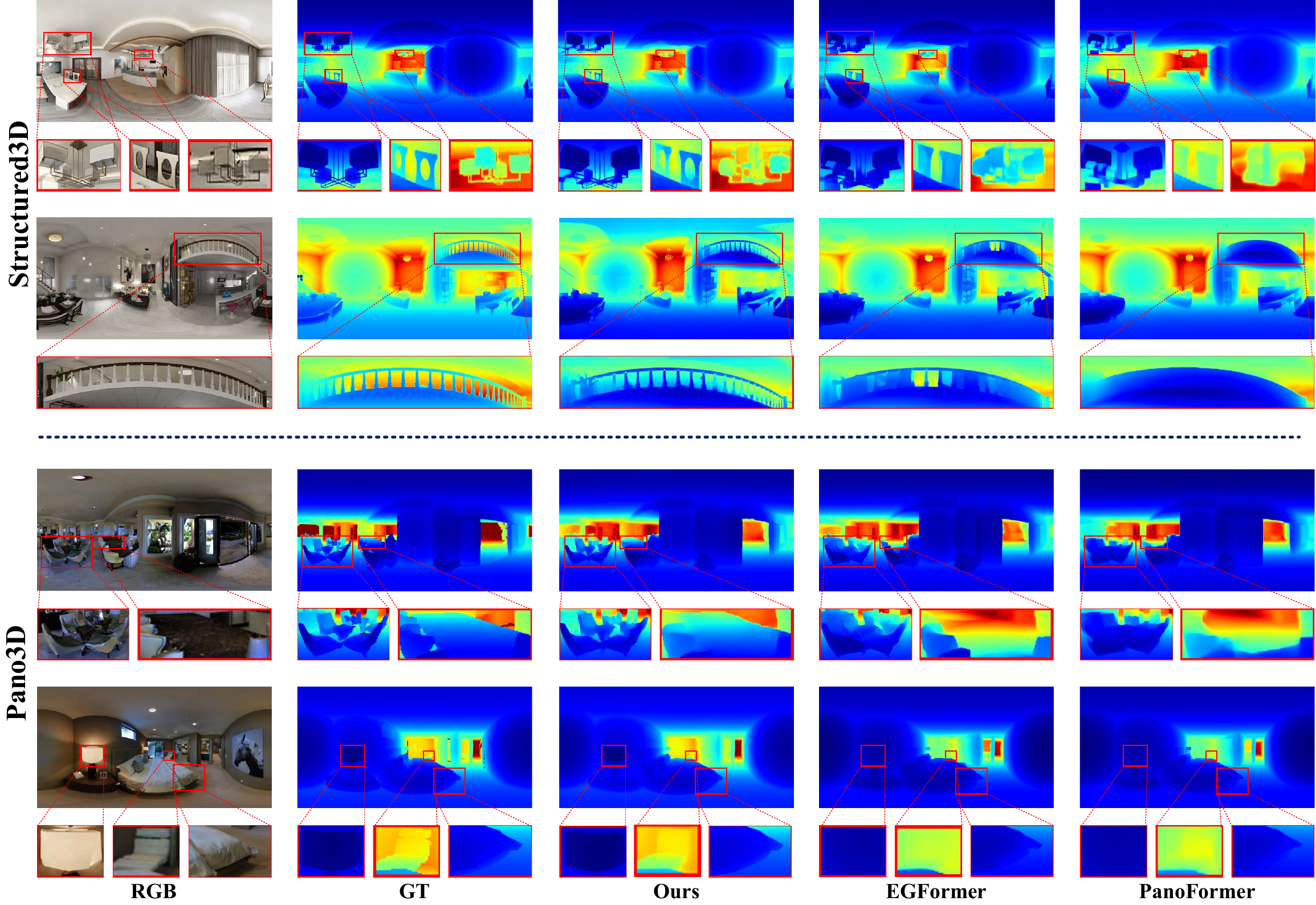}
\caption{\textbf { Qualitative results on the Structured3D dataset (the top two lines), and Pano3D dataset (the bottom two lines).}}
\label{fig9}
\end{figure*}

\section{Experiments}
We follow the training and evaluation protocol of EGformer \cite{r7} to evaluate our proposed SGFormer approach with two solid benchmarks. The implementation details are described as follows.  
\subsection{Dataset and Implementation Details}
Two datasets are employed in our evaluation: Structured3D \cite{r9} and Pano3D \cite{r10}. Structured3D is a high-quality synthetic dataset designed specifically to support a variety of structured 3D modeling tasks, providing many realistic images with detailed 3D structure annotations. On the other hand, the Pano3D dataset serves as a new benchmark for spherical panorama depth estimation. It aims to comprehensively evaluate the performance of depth prediction models, including the accuracy, the clarity of edges, and the image's smoothness. Consistent with EGFormer \cite{r7}, we follow the official dataset split \cite{r9,r10}. For the Structured3D dataset, we use scene\_00000 to scene\_02999 for training, scene\_03000 to scene\_03249 for validation, and scene\_3250 to scene\_03499 for testing. For the Pano3D dataset, we use 5636 images for training, 744 images for validation, and 1527 images for testing.

The reasons for not using PanoSunCG and 3D60 datasets are as follows: Firstly, the PanoSunCG dataset \cite{r31} is no longer publicly available. Secondly, the images in the 3D60 dataset \cite{r32} exhibit depth value leakage through their pixel brightness.


In our experimental settings, we perform all experiments using a single GTX 3090 GPU, where the batch size is set to 1. We choose the AdamW \cite{r35} as the optimizer. On the Structured3D dataset, the initial learning rate is set at $5e^{-5}$, with the model being trained for 50 epochs. Subsequently, after merging the Pano3D and Structured3D datasets, the model continues to be further trained for 20 epochs, maintaining the same learning rate and adopting a learning rate exponential decay strategy of 0.95 to optimize the training effect.

\begin{table*}[htbp]\centering
\setlength{\tabcolsep}{10pt}
\caption{ \centering Quantitative comparison on the S2D3D (Stanford2D3D) and M3D (Matterport3D) datasets with the SoTA methods. {\color{blue}\textbf{Blue}} indicates that our method achieves the best performance. $^{\dagger }$ means that we modify the structure of the HRDFuse network to ensure a fair comparison.
}
\label{table4}

{
\begin{tabular}{c c c|c c c c|c c c}
\hline
\textbf{Dataset} & \textbf{Method} & \textbf{Pub'Year} &\cellcolor{orange} \textbf{Abs.rel $\downarrow$}  & \cellcolor{orange}\textbf{Sq.rel $\downarrow$} & \cellcolor{orange}\textbf{RMSE $\downarrow$} & \cellcolor{orange}\textbf{MAE $\downarrow$} &\cellcolor{Ocean} \textbf{  $\delta ^1$ $\uparrow$} &\cellcolor{Ocean} \textbf{ $\delta^2$ $\uparrow$} & \cellcolor{Ocean}\textbf{  $\delta^3$ $\uparrow$} \\
\hline
            
            & EGFormer  \cite{r7} &ICCV'23    &0.1528 &0.1408 &0.4974 &0.2803 &0.8185 &0.9338 &0.9736 \\
            & PanoFormer  \cite{r6} &ECCV'22  & 0.1122 & 0.0786 & 0.3945  &0.2119 & 0.8874 & 0.9584 & 0.9859\\
            & OmniFusion   \cite{r3} &CVPR'22 &0.1154 &0.0775 &0.3809  &0.2164 &0.8674 &0.9603 &0.9871\\
S2D3D  \cite{r34}  & UniFuse  \cite{r2} &RAL'21 &0.1124 &0.0709 &0.3555  &\textbf{0.1978} &0.8706 &\textbf{0.9704} &0.9899\\
            & Elite360D  \cite{r45} &CVPR'24   &0.1182 &0.0728 &0.3756 &0.2219 &0.8872 &0.9684 &0.9892\\
            & SGFormer (ours) &-     &{\color{blue}\textbf{0.1040}} &{\color{blue}\textbf{0.0581}}  &{\color{blue}\textbf{0.3406}} &0.2017 &{\color{blue}\textbf{0.8998}} & {{0.9693}} & {\color{blue}\textbf{0.9908}}\\
\hline 
\hline 
                & EGFormer  \cite{r7} &ICCV'23    &0.1473 &0.1517 &0.6025 &0.3589 &0.8158 &0.9390 &0.9735 \\
            & PanoFormer  \cite{r6} &ECCV'22  & 0.1051 & 0.0966 & 0.4929  &0.2777 & 0.8908 & 0.9623 & 0.9831\\
            & BiFuse   \cite{r1} &CVPR'20 &0.1126 &0.0992 &0.5027  &0.2918 &0.8800 &0.9613 &0.9847\\
            & BiFuse++   \cite{r52} &TPAMI'22 &0.1123 &0.0915 &0.4853  &0.2783 & 0.8812 &0.9656 &0.9869\\
M3D  \cite{r33}  & UniFuse  \cite{r2} &RAL'21 &0.1144 &0.0936 &0.4835  &0.2821 &0.8785 &\textbf{0.9659} &0.9873\\
            & OmniFusion   \cite{r3} &CVPR'22 &0.1161 &0.1007 &0.4931  &0.2870 &0.8772 &0.9615 &0.9844\\
            & HRDFUSE$^{\dagger }$    \cite{r4} &CVPR'23 &0.1172 &0.0971 &0.5025 &0.2963 &0.8674 &0.9617 &0.9849\\
            & Elite360D  \cite{r45} &CVPR'24   &0.1115 &0.0914 &0.4875 &0.2826 &0.8815 &0.9646 &\textbf{0.9874}\\
            & SGFormer (ours) &-     &{\color{blue}\textbf{0.1039}} &{\color{blue}\textbf{0.0865}}  &{\color{blue}\textbf{0.4790}} &{\color{blue}\textbf{0.2748}} &{\color{blue}\textbf{0.8946}} & {{0.9642}} & {{0.9859}}\\
\hline 
\end{tabular}
}
\end{table*}

\subsection{Comparison Results}
\textbf{Metrics:}
To evaluate the model performance fairly, we select a series of standard evaluation metrics in EGFormer  \cite{r7}, including absolute relative error (Abs.rel), squared relative error (Sq.rel), root mean square linear error (RMS.lin), and root mean square log error (RMSlog). The lower the value of these indicators, the better the pmodel's performance. Additionally, we use relative accuracy ($\delta ^{n}$) as a measure of performance, where n represents different threshold levels (\textit{i.e.}, $\delta < 1.25^{n} $). The higher the value of this indicator, the higher the accuracy of the predictions.

\textbf{Quantitative Analysis:}
To validate the effectiveness of our approach, we compare our SGFormer model with the current state-of-the-art methods, including strategies for bi-projection fusion and long-range dependencies. Table \ref{table1} shows that our model ranks first in nearly all evaluation metrics, indicating superior overall performance. Specifically, on the one hand, compared with Bifuse \cite{r1} based on bi-projection fusion, our model has achieved a 53\% improvement (Abs.rel) and 41\% (RMS.lin) on the Structured3D dataset. For the Pano3D dataset, the performance is significantly improved by 66\% (Abs.rel) and 51\% (RMS.lin).
On the other hand, compared with EGFormer \cite{r7} based on long-range dependencies, our model achieves an 11\% improvement (Abs.rel) and 12\% (RMS.lin) on the Structured3D dataset. For the Pano3D dataset, SGFormer achieves 12\% (Abs.rel) and 9\% (RMS.lin) gain.
In quantitative comparison, we achieve significant performance improvements because we fully leverage the geometric priors of the sphere to relieve the panoramic distortion.
The lower quality of depth estimation in SliceNet \cite{r5} is mainly due to the detail loss when processing equirectangular images in the vertical direction.

\textbf{Qualitative Analysis:}
In Fig. \ref{fig9}, we present a qualitative comparison with EGFormer \cite{r7} and PanoFormer \cite{r6}. We can observe that our performance is superior in terms of regional structural details. Specifically, as demonstrated in the red boxes in Fig. \ref{fig9},  more accurate depths are recovered. We perform better on local details because we provide additional spherical geometric structures.

\textbf{More Comparison Results:}
We also compare our method with the latest method (CVPR'24) on additional datasets, despite the limitations of the Stanford2D3D \cite{r34} and Matterport3D \cite{r33} datasets. For instance, the Pano3D dataset \cite{r10} incorporates a rendered version of Matterport3D, making the use of both Pano3D and Matterport3D for evaluation redundant. Furthermore, in the Stanford2D3D dataset, the top and bottom parts of equirectangular images are not rendered properly, which handicaps our method when tested on them.
We strictly follow the Elite360D \cite{r45}. The quantitative comparison is shown in Table \ref{table4}. We achieve better performance on the Stanford2D3D dataset \cite{r34} and Matterport3D dataset \cite{r33}. 
Our dataset split is consistent with Elite360D \cite{r45}, following the official split \cite{r34,r33}.  For the Matterport3D dataset, we take 61 rooms for training and the others for testing. For the Stanford2D3D dataset, we take the fifth area (area 5) for testing, and the others are for training.





\begin{table*}[ht]\centering
\setlength{\tabcolsep}{10pt}
\caption{ \centering Ablation studies of individual components. We begin with a baseline model, adding each component sequentially. We conduct a series of ablation studies on the S3D (Structured3D) and Pano3D datasets. Results in bold indicate the best results. (CLE: curve local embedding, GCPE: global conditional position embedding, CR: circular rotation, BRP: bipolar re-projection) }
\label{table2}
{
\begin{tabular}{ c|c|c|c|c|c|c|c|c|c|c}
\hline
  Data & ID & CLE & GCPE & CR &BRP & \cellcolor{orange}\textbf{Abs.rel $\downarrow$}  & \cellcolor{orange}\textbf{Sq.rel $\downarrow$} & \cellcolor{orange}\textbf{RMS.lin $\downarrow$}  & \cellcolor{orange}\textbf{RMSlog $\downarrow$} &\cellcolor{Ocean} \textbf{  $\delta ^1$ $\uparrow$}  \\
\hline
&a) &\XSolidBrush &\XSolidBrush &\XSolidBrush &\XSolidBrush &0.0388 &0.0352 &0.2838  &0.0968  &0.9800  \\
&b) &\Checkmark &\XSolidBrush &\XSolidBrush &\XSolidBrush &0.0362 &0.0333 &0.2652  &0.0924  &0.9822  \\
S3D \cite{r9} &c) &\Checkmark &\Checkmark &\XSolidBrush &\XSolidBrush 
&0.0329  &0.0307  &0.2561  &0.0898  &0.9841  \\
&d) &\Checkmark &\Checkmark &\Checkmark & \XSolidBrush 
&\textbf{0.0301}  &0.0299  &0.2499  &0.0874 &0.9849 \\
&e) &\Checkmark &\Checkmark &\Checkmark &\Checkmark &0.0303 &\textbf{0.0293} &\textbf{0.2429}  &\textbf{0.0863} &\textbf{0.9857} \\
\hline
\hline
&a) &\XSolidBrush &\XSolidBrush &\XSolidBrush &\XSolidBrush &0.0720 &0.0483 &0.4048  &0.1278  &0.9417 \\
&b) &\Checkmark &\XSolidBrush &\XSolidBrush &\XSolidBrush &0.0708 &0.0436 &0.3814 &0.1238  &0.9474  \\
Pano3D  \cite{r10} &c) &\Checkmark &\Checkmark &\XSolidBrush &\XSolidBrush
&0.0649  &0.0416  &0.3774  &0.1183  &0.9543  \\
&d) &\Checkmark &\Checkmark &\Checkmark & \XSolidBrush 
&0.0626  &0.0382  &0.3619  &0.1117 &0.9598\\
&e) &\Checkmark &\Checkmark &\Checkmark &\Checkmark & \textbf{0.0583} & \textbf{0.0364} & \textbf{0.3537}  &\textbf{0.1101} &\textbf{0.9613}\\
\hline
\hline
\end{tabular}
}
\end{table*}

\begin{figure*}[ht]
\centering
\includegraphics[scale=0.55]{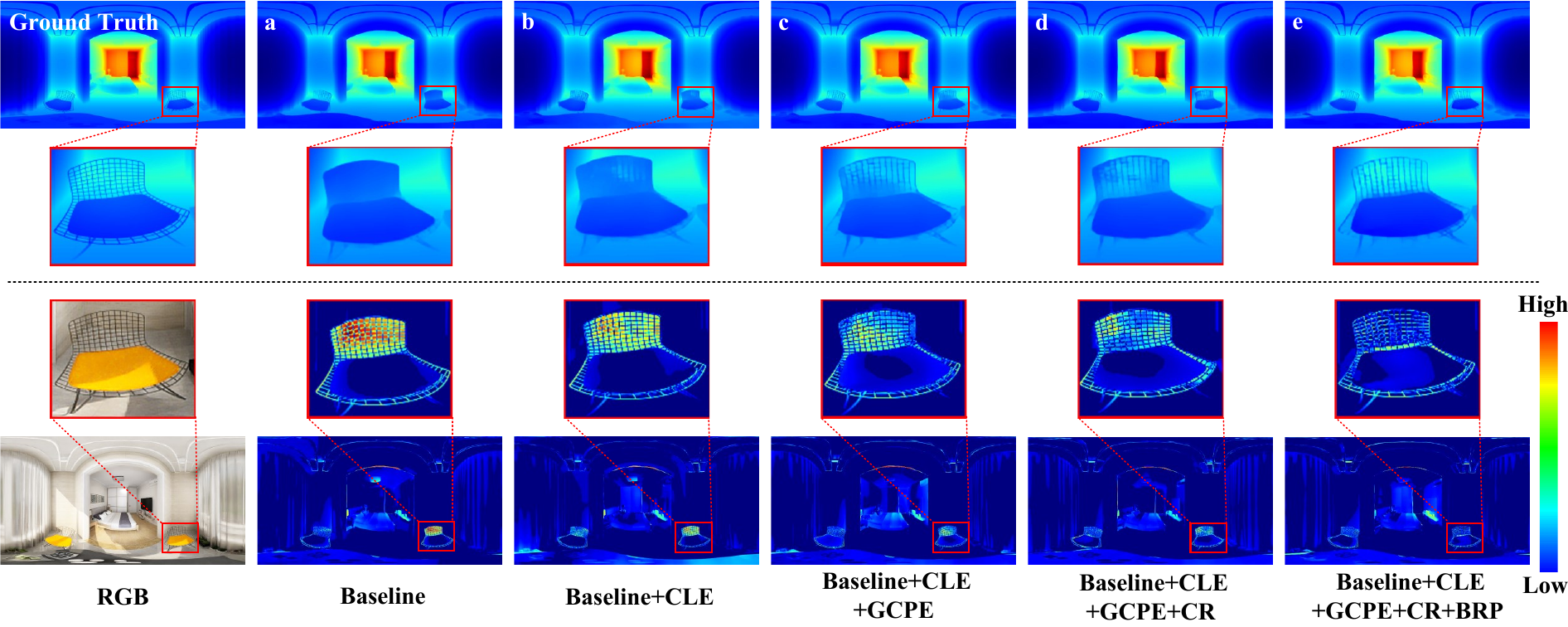}
\caption{\textbf {Qualitative comparisons about individual components. The top row shows visual comparisons of depth estimation after adding different components, while the bottom row presents the visual comparisons of the error maps corresponding to the estimated depth maps. The middle two rows display close-up views of the highlighted areas in the top and bottom rows, respectively. As we add more components from left to right, the object boundaries become clearer, and the overall depth estimation becomes more accurate with lower errors. The trend of error changes can be directly observed from the error maps. a, b, c, d, e are the same as Table \ref{table2}.}}
\label{fig10}
\end{figure*}

\begin{figure}[ht]
\centering
\includegraphics[scale=0.38]{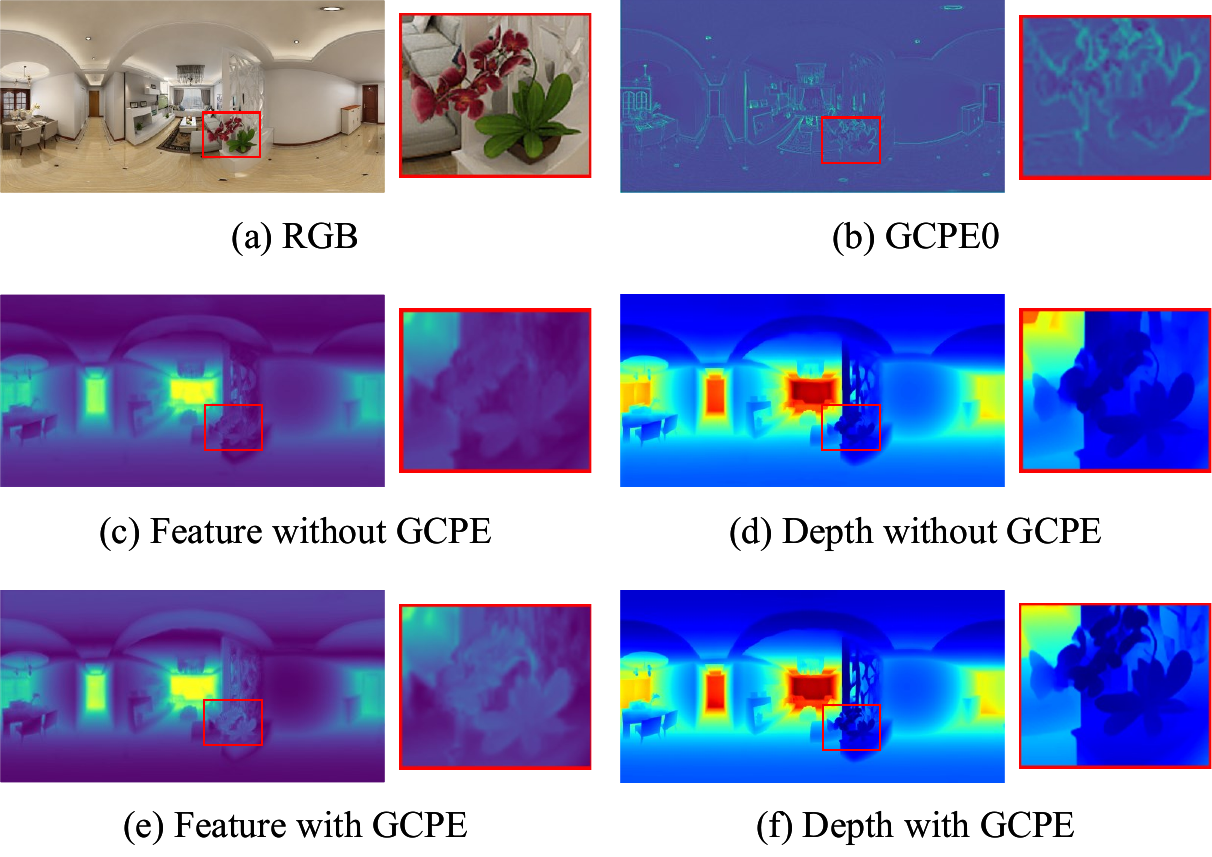}
\caption{\textbf {The visualization of heatmap of features and  the predicted depth map with/without GCPE.}}
\label{fig4}
\end{figure}

\subsection{Ablation study}
We thoroughly validate the key components of the SGFormer model under the same experimental conditions. As shown in Table \ref{table2}, initially, we establish a baseline with a model that does not include any specific components, using a Resnet-34 as the encoder and a ViT-based \cite{r26} decoder. Then, we sequentially introduce core components on the Structured3D and Pano3D datasets to explore the effectiveness of our model.

\subsubsection{Effectiveness of curve local embedding (CLE)}
From Table \ref{table2}, we can observe that on the Structured3D dataset, there is a 6.70\% improvement (Abs.rel) and 6.55\% (RMS.lin). On the Pano3D dataset, the performance is further boosted by 1.67\% (Abs Rel) and 5.78\% (RMS.lin). These results illustrate that by incorporating CLE, we introduce additional spherical surface distance, which aids the network in perceiving distortion distribution.   

\subsubsection{Effectiveness of global conditional position embedding (GCPE)}
The introduction of GCPE compensates for the model’s geometric details and spatial structure across different resolutions. Results in Table \ref{table2} show that the performance is significantly improved by 9.12\% (Abs.rel) and 3.43\% (RMS.lin) on the structured3D dataset. For the Pano3D dataset, there is an improvement of 8.33\%  (Abs.rel) and 1.05\% (RMS.lin).

\subsubsection{Effectiveness of circular rotation (CR)}
The CR component is added to achieve cross-window feature interaction. Comparisons in Table \ref{table2} reveal an 8.51\% improvement (Abs.rel) and 2.42\% (RMS.lin) on the Structured3D dataset. The performance is further enhanced by 3.54\%  (Abs.rel) and 4.11\%  (RMS.lin) on the Pano3D dataset.

\subsubsection{Effectiveness of bipolar re-projection (BRP)}
The BRP component is incorporated to reduce distortion by projecting the north and south polar regions near the equator. Results from Table \ref{table2} show that performance is improved by 2.80\% (RMS.lin) on the Structured3D dataset. For the Pano3D dataset, the performance is further boosted by 6.87\% (Abs.rel) and 2.27\% (RMS.lin).

In Fig. \ref{fig10}, as we gradually add CLE, GCPE, CR, and BRP components from left to right, the output depth maps perform better in overall depth estimation and capture local structural details more precisely. 
Furthermore, through the visualization comparison of error maps, we can see that the estimated error is continuously decreasing (Darker shades of the red indicate a higher error, while darker shades of the blue signify a lower error.).

\subsubsection{GCPE}From Fig. \ref{fig4}, the depth details can be observed clearly, indicating that the query-based GCPE strategy can contribute to the spatial structure and depth details.

\subsubsection{Backbones}
Following the previous works\cite{r3,r45}, we use ResNet-34 to extract features from the ERP images. In addition, we also replace the backbone (ResNet-34) with a more advanced transformer-based architecture (\textit{i.e.}, Swin-B) to verify the extensibility of our proposed SPDecoder. As shown in Table \ref{table3}, the combination of the advanced backbone and our designs can obtain a significant improvement across all metrics. This improvement indicates that although the traditional self-attention mechanism still can not address the distortion issue \cite{r6}, our distortion-aware designs can fully use the advantages of these advanced backbones' large receptive field, delivering superior performance. Considering the computational cost, we employ ResNet-34 as the backbone to extract ERP features.

\begin{table}[H]
\centering
\caption{ Results of using different backbones }
\scalebox{0.80}
{\begin{tabular}{ c|c|c|c|c|c|c}
\hline
  Dataset & Backbone & \cellcolor{orange}\textbf{Abs.rel $\downarrow$}  & \cellcolor{orange}\textbf{Sq.rel $\downarrow$} & \cellcolor{orange}\textbf{RMS.lin $\downarrow$}  & \cellcolor{orange}\textbf{RMSlog $\downarrow$} &\cellcolor{Ocean} \textbf{  $\delta ^1$ $\uparrow$}  \\
\hline
\multirow{2}*{Pano3D \cite{r10}} &ResNet-34 &0.0583 &0.0364 &0.3537  &0.1101  &0.9613  \\
&Swin-B &\textbf{0.0519} &\textbf{0.0270} &\textbf{0.3066}  &\textbf{0.0953}  &\textbf{0.9725}  \\
\hline
\end{tabular}}
\label{table3}
\end{table}

\section{Conclusion}
In this paper, we propose a spherical geometry transformer, SGFormer, integrating three geometric priors: equidistortion, continuity, and surface distance, to mitigate distortion effects. 
We employ bipolar re-projection (BRP) to project polar regions towards the equator.
This strategy is parameter-free and efficient without additional branches and complex transformation costs.
Circular rotation (CR) leverages sphere continuity to enhance cross-window feature interaction and understanding of global structure. 
Curve local embedding (CLE) uses relative position embeddings from pixels projected onto spherical space, explicitly perceiving distortion with additional surface distance. 
Additionally, we introduce a query-based global conditional position embedding strategy, which can dynamically provide varying spatial structure and geometric local details across different resolutions.
Experiments on two benchmarks demonstrate that SGFormer significantly outperforms SoTA methods.

\bibliographystyle{IEEEtran}
\bibliography{ref}

\begin{thebibliography}{10}
\providecommand{\url}[1]{#1}
\csname url@samestyle\endcsname
\providecommand{\newblock}{\relax}
\providecommand{\bibinfo}[2]{#2}
\providecommand{\BIBentrySTDinterwordspacing}{\spaceskip=0pt\relax}
\providecommand{\BIBentryALTinterwordstretchfactor}{4}
\providecommand{\BIBentryALTinterwordspacing}{\spaceskip=\fontdimen2\font plus
\BIBentryALTinterwordstretchfactor\fontdimen3\font minus \fontdimen4\font\relax}
\providecommand{\BIBforeignlanguage}[2]{{%
\expandafter\ifx\csname l@#1\endcsname\relax
\typeout{** WARNING: IEEEtran.bst: No hyphenation pattern has been}%
\typeout{** loaded for the language `#1'. Using the pattern for}%
\typeout{** the default language instead.}%
\else
\language=\csname l@#1\endcsname
\fi
#2}}
\providecommand{\BIBdecl}{\relax}
\BIBdecl

\bibitem{r1}
F.-E. Wang, Y.-H. Yeh, M.~Sun, W.-C. Chiu, and Y.-H. Tsai, ``Bifuse: Monocular 360 depth estimation via bi-projection fusion,'' in \emph{Proceedings of the IEEE/CVF Conference on Computer Vision and Pattern Recognition}, 2020, pp. 462--471.

\bibitem{r7}
I.~Yun, C.~Shin, H.~Lee, H.-J. Lee, and C.~E. Rhee, ``Egformer: Equirectangular geometry-biased transformer for 360 depth estimation,'' in \emph{Proceedings of the IEEE/CVF International Conference on Computer Vision}, 2023, pp. 6101--6112.

\bibitem{r46}
L.~Papa, P.~Russo, and I.~Amerini, ``Meter: A mobile vision transformer architecture for monocular depth estimation,'' \emph{IEEE Transactions on Circuits and Systems for Video Technology}, vol.~33, no.~10, pp. 5882--5893, 2023.

\bibitem{r47}
------, ``D4d: An rgbd diffusion model to boost monocular depth estimation,'' \emph{IEEE Transactions on Circuits and Systems for Video Technology}, 2024.

\bibitem{r48}
H.~Kumar, A.~S. Yadav, S.~Gupta, and K.~S. Venkatesh, ``Depth map estimation using defocus and motion cues,'' \emph{IEEE Transactions on Circuits and Systems for Video Technology}, vol.~29, no.~5, pp. 1365--1379, 2019.

\bibitem{r49}
X.~Meng, C.~Fan, Y.~Ming, and H.~Yu, ``Cornet: Context-based ordinal regression network for monocular depth estimation,'' \emph{IEEE Transactions on Circuits and Systems for Video Technology}, vol.~32, no.~7, pp. 4841--4853, 2022.

\bibitem{r50}
Y.~Cao, T.~Zhao, K.~Xian, C.~Shen, Z.~Cao, and S.~Xu, ``Monocular depth estimation with augmented ordinal depth relationships,'' \emph{IEEE Transactions on Circuits and Systems for Video Technology}, vol.~30, no.~8, pp. 2674--2682, 2020.

\bibitem{r51}
H.~Yan, X.~Yu, Y.~Zhang, S.~Zhang, X.~Zhao, and L.~Zhang, ``Single image depth estimation with normal guided scale invariant deep convolutional fields,'' \emph{IEEE Transactions on Circuits and Systems for Video Technology}, vol.~29, no.~1, pp. 80--92, 2019.

\bibitem{r53}
R.~Cong, C.~Wu, X.~Song, W.~Zhang, S.~Kwong, H.~Li, and P.~Ji, ``Srnsd: Structure-regularized night-time self-supervised monocular depth estimation for outdoor scenes,'' \emph{IEEE Transactions on Image Processing}, 2024.

\bibitem{r4}
H.~Ai, Z.~Cao, Y.-P. Cao, Y.~Shan, and L.~Wang, ``Hrdfuse: Monocular 360deg depth estimation by collaboratively learning holistic-with-regional depth distributions,'' in \emph{Proceedings of the IEEE/CVF Conference on Computer Vision and Pattern Recognition}, 2023, pp. 13\,273--13\,282.

\bibitem{r2}
H.~Jiang, Z.~Sheng, S.~Zhu, Z.~Dong, and R.~Huang, ``Unifuse: Unidirectional fusion for 360 panorama depth estimation,'' \emph{IEEE Robotics and Automation Letters}, vol.~6, no.~2, pp. 1519--1526, 2021.

\bibitem{r3}
Y.~Li, Y.~Guo, Z.~Yan, X.~Huang, Y.~Duan, and L.~Ren, ``Omnifusion: 360 monocular depth estimation via geometry-aware fusion,'' in \emph{Proceedings of the IEEE/CVF Conference on Computer Vision and Pattern Recognition}, 2022, pp. 2801--2810.

\bibitem{r5}
G.~Pintore, M.~Agus, E.~Almansa, J.~Schneider, and E.~Gobbetti, ``Slicenet: deep dense depth estimation from a single indoor panorama using a slice-based representation,'' in \emph{Proceedings of the IEEE/CVF Conference on Computer Vision and Pattern Recognition}, 2021, pp. 11\,536--11\,545.

\bibitem{r6}
Z.~Shen, C.~Lin, K.~Liao, L.~Nie, Z.~Zheng, and Y.~Zhao, ``Panoformer: Panorama transformer for indoor 360 depth estimation,'' in \emph{European Conference on Computer Vision}.\hskip 1em plus 0.5em minus 0.4em\relax Springer, 2022, pp. 195--211.

\bibitem{r36}
S.~Hochreiter and J.~Schmidhuber, ``Long short-term memory,'' \emph{Neural computation}, vol.~9, no.~8, pp. 1735--1780, 1997.

\bibitem{r9}
J.~Zheng, J.~Zhang, J.~Li, R.~Tang, S.~Gao, and Z.~Zhou, ``Structured3d: A large photo-realistic dataset for structured 3d modeling,'' in \emph{Computer Vision--ECCV 2020: 16th European Conference, Glasgow, UK, August 23--28, 2020, Proceedings, Part IX 16}.\hskip 1em plus 0.5em minus 0.4em\relax Springer, 2020, pp. 519--535.

\bibitem{r10}
G.~Albanis, N.~Zioulis, P.~Drakoulis, V.~Gkitsas, V.~Sterzentsenko, F.~Alvarez, D.~Zarpalas, and P.~Daras, ``Pano3d: A holistic benchmark and a solid baseline for 360deg depth estimation,'' in \emph{Proceedings of the IEEE/CVF Conference on Computer Vision and Pattern Recognition}, 2021, pp. 3727--3737.

\bibitem{r11}
J.~Bai, S.~Lai, H.~Qin, J.~Guo, and Y.~Guo, ``Glpanodepth: Global-to-local panoramic depth estimation,'' \emph{arXiv preprint arXiv:2202.02796}, 2022.

\bibitem{r12}
M.~Rey-Area, M.~Yuan, and C.~Richardt, ``360monodepth: High-resolution 360deg monocular depth estimation,'' in \emph{Proceedings of the IEEE/CVF Conference on Computer Vision and Pattern Recognition}, 2022, pp. 3762--3772.

\bibitem{r67}
C.-H. Peng and J.~Zhang, ``High-resolution depth estimation for 360deg panoramas through perspective and panoramic depth images registration,'' in \emph{Proceedings of the IEEE/CVF Winter Conference on Applications of Computer Vision}, 2023, pp. 3116--3125.

\bibitem{r13}
B.~Coors, A.~P. Condurache, and A.~Geiger, ``Spherenet: Learning spherical representations for detection and classification in omnidirectional images,'' in \emph{Proceedings of the European conference on computer vision (ECCV)}, 2018, pp. 518--533.

\bibitem{r14}
J.~Dai, H.~Qi, Y.~Xiong, Y.~Li, G.~Zhang, H.~Hu, and Y.~Wei, ``Deformable convolutional networks,'' in \emph{Proceedings of the IEEE international conference on computer vision}, 2017, pp. 764--773.

\bibitem{r17}
X.~Zhu, W.~Su, L.~Lu, B.~Li, X.~Wang, and J.~Dai, ``Deformable detr: Deformable transformers for end-to-end object detection,'' \emph{arXiv preprint arXiv:2010.04159}, 2020.

\bibitem{r18}
B.~Xiong and K.~Grauman, ``Snap angle prediction for 360 panoramas,'' in \emph{Proceedings of the European Conference on Computer Vision (ECCV)}, 2018, pp. 3--18.

\bibitem{r20}
C.~Jiang, J.~Huang, K.~Kashinath, P.~Marcus, M.~Niessner \emph{et~al.}, ``Spherical cnns on unstructured grids,'' \emph{arXiv preprint arXiv:1901.02039}, 2019.

\bibitem{r21}
R.~Khasanova and P.~Frossard, ``Geometry aware convolutional filters for omnidirectional images representation,'' in \emph{International Conference on Machine Learning}.\hskip 1em plus 0.5em minus 0.4em\relax PMLR, 2019, pp. 3351--3359.

\bibitem{r22}
I.~Laina, C.~Rupprecht, V.~Belagiannis, F.~Tombari, and N.~Navab, ``Deeper depth prediction with fully convolutional residual networks,'' in \emph{2016 Fourth international conference on 3D vision (3DV)}.\hskip 1em plus 0.5em minus 0.4em\relax IEEE, 2016, pp. 239--248.

\bibitem{r16}
Y.-C. Su and K.~Grauman, ``Kernel transformer networks for compact spherical convolution,'' in \emph{Proceedings of the IEEE/CVF Conference on Computer Vision and Pattern Recognition}, 2019, pp. 9442--9451.

\bibitem{r23}
T.~S. Cohen, M.~Geiger, J.~K{\"o}hler, and M.~Welling, ``Spherical cnns,'' \emph{arXiv preprint arXiv:1801.10130}, 2018.

\bibitem{r25}
Z.~Shen, C.~Lin, L.~Nie, K.~Liao, and Y.~Zhao, ``Neural contourlet network for monocular 360 depth estimation,'' \emph{IEEE Transactions on Circuits and Systems for Video Technology}, vol.~32, no.~12, pp. 8574--8585, 2022.

\bibitem{r72}
N.-H. Wang and Y.-L. Liu, ``Depth anywhere: Enhancing 360 monocular depth estimation via perspective distillation and unlabeled data augmentation,'' \emph{arXiv preprint arXiv:2406.12849}, 2024.

\bibitem{r75}
L.~Yang, B.~Kang, Z.~Huang, X.~Xu, J.~Feng, and H.~Zhao, ``Depth anything: Unleashing the power of large-scale unlabeled data,'' in \emph{Proceedings of the IEEE/CVF Conference on Computer Vision and Pattern Recognition}, 2024, pp. 10\,371--10\,381.

\bibitem{r74}
Q.~Feng, H.~P. Shum, and S.~Morishima, ``360 depth estimation in the wild-the depth360 dataset and the segfuse network,'' in \emph{2022 IEEE Conference on Virtual Reality and 3D User Interfaces (VR)}.\hskip 1em plus 0.5em minus 0.4em\relax IEEE, 2022, pp. 664--673.

\bibitem{r70}
Y.~Zhang, L.~Chu, Z.~Wang, H.~Tong, J.~Hu, and J.~Li, ``A novel panorama depth estimation framework for autonomous driving scenarios based on a vision transformer,'' \emph{Sensors}, vol.~24, no.~21, p. 7013, 2024.

\bibitem{r73}
M.~Sch{\"o}n, J.~Ruof, T.~Wodtko, M.~Buchholz, and K.~Dietmayer, ``The aduulm-360 dataset--a multi-modal dataset for depth estimation in adverse weather,'' \emph{arXiv preprint arXiv:2411.11455}, 2024.

\bibitem{r26}
A.~Dosovitskiy, L.~Beyer, A.~Kolesnikov, D.~Weissenborn, X.~Zhai, T.~Unterthiner, M.~Dehghani, M.~Minderer, G.~Heigold, S.~Gelly \emph{et~al.}, ``An image is worth 16x16 words: Transformers for image recognition at scale,'' \emph{arXiv preprint arXiv:2010.11929}, 2020.

\bibitem{r27}
Z.~Liu, Y.~Lin, Y.~Cao, H.~Hu, Y.~Wei, Z.~Zhang, S.~Lin, and B.~Guo, ``Swin transformer: Hierarchical vision transformer using shifted windows,'' in \emph{Proceedings of the IEEE/CVF international conference on computer vision}, 2021, pp. 10\,012--10\,022.

\bibitem{r71}
D.~Shim and H.~J. Kim, ``Swindepth: Unsupervised depth estimation using monocular sequences via swin transformer and densely cascaded network,'' in \emph{2023 IEEE International Conference on Robotics and Automation (ICRA)}.\hskip 1em plus 0.5em minus 0.4em\relax IEEE, 2023, pp. 4983--4990.

\bibitem{r8}
X.~Dong, J.~Bao, D.~Chen, W.~Zhang, N.~Yu, L.~Yuan, D.~Chen, and B.~Guo, ``Cswin transformer: A general vision transformer backbone with cross-shaped windows,'' in \emph{Proceedings of the IEEE/CVF conference on computer vision and pattern recognition}, 2022, pp. 12\,124--12\,134.

\bibitem{r28}
Z.~Ling, Z.~Xing, X.~Zhou, M.~Cao, and G.~Zhou, ``Panoswin: a pano-style swin transformer for panorama understanding,'' in \emph{Proceedings of the IEEE/CVF Conference on Computer Vision and Pattern Recognition}, 2023, pp. 17\,755--17\,764.

\bibitem{r29}
S.~Cho, R.~Jung, and J.~Kwon, ``Spherical transformer,'' \emph{arXiv preprint arXiv:2202.04942}, 2022.

\bibitem{r68}
X.~Huang, Y.~Ma, Z.~Yu, and H.~Zhao, ``Rcdformer: Transformer-based dense depth estimation by sparse radar and camera,'' \emph{Neurocomputing}, vol. 589, p. 127668, 2024.

\bibitem{r54}
R.~Cong, K.~Huang, J.~Lei, Y.~Zhao, Q.~Huang, and S.~Kwong, ``Multi-projection fusion and refinement network for salient object detection in 360 omnidirectional image,'' \emph{IEEE Transactions on Neural Networks and Learning Systems}, 2023.

\bibitem{r57}
H.-T. Cheng, C.-H. Chao, J.-D. Dong, H.-K. Wen, T.-L. Liu, and M.~Sun, ``Cube padding for weakly-supervised saliency prediction in 360 videos,'' in \emph{Proceedings of the IEEE conference on computer vision and pattern recognition}, 2018, pp. 1420--1429.

\bibitem{r58}
Z.~Zhang, Y.~Xu, J.~Yu, and S.~Gao, ``Saliency detection in 360 videos,'' in \emph{Proceedings of the European conference on computer vision (ECCV)}, 2018, pp. 488--503.

\bibitem{r69}
Y.~Zhao, L.~Zhao, Q.~Yu, L.~Sheng, J.~Zhang, and D.~Xu, ``Distortion-aware transformer in 360 salient object detection,'' in \emph{Proceedings of the 31st ACM International Conference on Multimedia}, 2023, pp. 499--508.

\bibitem{r59}
Q.~Zhao, C.~Zhu, F.~Dai, Y.~Ma, G.~Jin, and Y.~Zhang, ``Distortion-aware cnns for spherical images.'' in \emph{IJCAI}, 2018, pp. 1198--1204.

\bibitem{r60}
Y.~Xu, Z.~Zhang, and S.~Gao, ``Spherical dnns and their applications in 360 images and videos,'' \emph{IEEE Transactions on Pattern Analysis and Machine Intelligence}, vol.~44, no.~10, pp. 7235--7252, 2021.

\bibitem{r65}
Y.~Guo, M.~Xu, L.~Jiang, L.~Sigal, and Y.~Chen, ``Dinn360: Deformable invertible neural network for latitude-aware 360deg image rescaling,'' in \emph{Proceedings of the IEEE/CVF Conference on Computer Vision and Pattern Recognition}, 2023, pp. 21\,519--21\,528.

\bibitem{r61}
Y.~Lee, J.~Jeong, J.~Yun, W.~Cho, and K.-J. Yoon, ``Spherephd: Applying cnns on a spherical polyhedron representation of 360deg images,'' in \emph{Proceedings of the IEEE/CVF Conference on Computer Vision and Pattern Recognition}, 2019, pp. 9181--9189.

\bibitem{r62}
M.~Eder, M.~Shvets, J.~Lim, and J.-M. Frahm, ``Tangent images for mitigating spherical distortion,'' in \emph{Proceedings of the IEEE/CVF Conference on Computer Vision and Pattern Recognition}, 2020, pp. 12\,426--12\,434.

\bibitem{r63}
H.~Singh, M.~Verma, and R.~Cheruku, ``Dsnet: Efficient lightweight model for video salient object detection for iot and wot applications,'' in \emph{Companion Proceedings of the ACM Web Conference 2023}, 2023, pp. 1286--1295.

\bibitem{r64}
------, ``Dsfnet: Video salient object detection using a novel lightweight deformable separable fusion network,'' \emph{IEEE Transactions on Instrumentation and Measurement}, 2024.

\bibitem{r37}
A.~Vaswani, N.~Shazeer, N.~Parmar, J.~Uszkoreit, L.~Jones, A.~N. Gomez, {\L}.~Kaiser, and I.~Polosukhin, ``Attention is all you need,'' \emph{Advances in neural information processing systems}, vol.~30, 2017.

\bibitem{r39}
R.~W. Sinnott, ``Virtues of the haversine,'' \emph{Sky and telescope}, vol.~68, no.~2, p. 158, 1984.

\bibitem{r30}
I.~Yun, H.-J. Lee, and C.~E. Rhee, ``Improving 360 monocular depth estimation via non-local dense prediction transformer and joint supervised and self-supervised learning,'' in \emph{Proceedings of the AAAI Conference on Artificial Intelligence}, vol.~36, no.~3, 2022, pp. 3224--3233.

\bibitem{r45}
H.~Ai and L.~Wang, ``Elite360d: Towards efficient 360 depth estimation via semantic-and distance-aware bi-projection fusion,'' in \emph{Proceedings of the IEEE/CVF Conference on Computer Vision and Pattern Recognition}, 2024, pp. 9926--9935.

\bibitem{r38}
R.~Ranftl, K.~Lasinger, D.~Hafner, K.~Schindler, and V.~Koltun, ``Towards robust monocular depth estimation: Mixing datasets for zero-shot cross-dataset transfer,'' \emph{IEEE transactions on pattern analysis and machine intelligence}, vol.~44, no.~3, pp. 1623--1637, 2020.

\bibitem{r31}
F.-E. Wang, H.-N. Hu, H.-T. Cheng, J.-T. Lin, S.-T. Yang, M.-L. Shih, H.-K. Chu, and M.~Sun, ``Self-supervised learning of depth and camera motion from 360 videos,'' in \emph{Asian Conference on Computer Vision}.\hskip 1em plus 0.5em minus 0.4em\relax Springer, 2018, pp. 53--68.

\bibitem{r32}
N.~Zioulis, A.~Karakottas, D.~Zarpalas, and P.~Daras, ``Omnidepth: Dense depth estimation for indoors spherical panoramas,'' in \emph{Proceedings of the European Conference on Computer Vision (ECCV)}, 2018, pp. 448--465.

\bibitem{r35}
I.~Loshchilov and F.~Hutter, ``Decoupled weight decay regularization,'' \emph{arXiv preprint arXiv:1711.05101}, 2017.

\bibitem{r34}
I.~Armeni, S.~Sax, A.~R. Zamir, and S.~Savarese, ``Joint 2d-3d-semantic data for indoor scene understanding,'' \emph{arXiv preprint arXiv:1702.01105}, 2017.

\bibitem{r52}
F.-E. Wang, Y.-H. Yeh, Y.-H. Tsai, W.-C. Chiu, and M.~Sun, ``Bifuse++: Self-supervised and efficient bi-projection fusion for 360 depth estimation,'' \emph{IEEE transactions on pattern analysis and machine intelligence}, vol.~45, no.~5, pp. 5448--5460, 2022.

\bibitem{r33}
A.~Chang, A.~Dai, T.~Funkhouser, M.~Halber, M.~Niessner, M.~Savva, S.~Song, A.~Zeng, and Y.~Zhang, ``Matterport3d: Learning from rgb-d data in indoor environments,'' \emph{arXiv preprint arXiv:1709.06158}, 2017.

\end{thebibliography}

\end{document}